\documentclass[11pt]{article}
\usepackage{coling2020}
\usepackage{times}
\usepackage{url}
\usepackage{latexsym}
\usepackage{amsfonts}
\usepackage{multirow}
\usepackage{amssymb}
\usepackage{subfigure}
\usepackage{graphicx}
\usepackage{arydshln}
\usepackage{amsmath}
\usepackage{caption}
\usepackage{CJKutf8}
\usepackage{caption}
\usepackage{capt-of}

\colingfinalcopy % Uncomment this line for the final submission

\title{Does Chinese BERT Encode Word Structure?}

\author{Yile Wang\textsuperscript{\rm 1,2,3}\textnormal{,}\ Leyang Cui\textsuperscript{\rm 1,2,3} \textnormal{and} Yue Zhang\textsuperscript{\rm 2,3} \\
 \textsuperscript{\rm 1}Zhejiang University \\
  \textsuperscript{\rm 2}School of Engineering, Westlake University \\
 \textsuperscript{\rm 3}Institute of Advanced Technology, Westlake Institute for Advanced Study \\
  \texttt{\{wangyile,cuileyang\}@westlake.edu.cn} \\
  \texttt{yue.zhang@wias.org.cn} \\
}

\date{}

\begin{document}
\maketitle
\begin{abstract}
Contextualized representations give significantly improved results for a wide range of NLP tasks. Much work has been dedicated to analyzing the features captured by representative models such as BERT. Existing work finds that syntactic, semantic and word sense knowledge are encoded in BERT. However, little work has investigated word features for character-based languages such as Chinese. We investigate Chinese BERT using both attention weight distribution statistics and probing tasks, finding that (1) word information is captured by BERT; (2) word-level features are mostly in the middle representation layers; (3) downstream tasks make different use of word features in BERT, with POS tagging and chunking relying the most on word features, and natural language inference relying the least on such features.
\end{abstract}

\section{Introduction}
Large scale pre-trained models such as BERT~\cite{bert} have been widely used in NLP, giving improved results in a wide range of downstream tasks, including dependency parsing~\cite{bert-dependency-parsing}, summarization~\cite{bert-summarization} and reading comprehension~\cite{bert-rc}. To better understand the reason behind their effectiveness, a natural question is what knowledge can be learned during the self-supervised pre-training stage. Some previous works prove that BERT effectively captures syntactic information~\cite{Goldberg2019AssessingBS}, semantic information~\cite{analysis-bert-1}, commonsense knowledge~\cite{commonsense}, and factual knowledge~\cite{petroni-etal-2019-language} without fine-tuning on task-specific datasets, which explains its effectiveness on related tasks.

BERT has also gained success in Chinese NLP~\cite{rc,sun2019ernie,zhang-etal-2019-ernie}. 
However, relatively little work investigate knowledge learned by Chinese BERT. 
%One of the major difference between Chinese and English is that the former are naturally written as sequences of characters without explicit word boundary. 
Different from English, Chinese sentences are written as sequences of characters without explicit word boundary. Yet Chinese BERT is character-based, where the contextualized representations are built on each character. 
%However, the existing Chinese pre-trained model is character-based, which means the contextualized representation in build on each character. 
For traditional Chinese NLP, word segmentation is considered an essential pre-processing step~\cite{cai-neural-seg,dependency-seg}.
%Moreover, previous work~\cite{lattice-lstm} demonstrates that leveraging word information is able to improve the representation power of character sequence encoder.
In neural modeling, Li et al.~\shortcite{wordseg} argue that character-based model consistently outperforms word-based model for Chinese, because the former can alleviate the over-fitting and out-of-vocabulary issue. 
% However, \cite{wordseg} find that character-based model consistently outperforms word-based model for Chinese. They argue that the former can alleviate the over-fitting and out-of-vocabulary issue.
% However, the existing Chinese pre-trained model is character-based, which means the contextualized representation in build on each character. 
%It is still unclear whether character-based pre-trained model solve Chinese tasks effectively.
In this paper, we investigate whether Chinese BERT encodes word structure features.%, thus supporting it success on downstream tasks.

We aim to answer the following three research questions. First, how much word information is captured by character-based Chinese BERT? Second, out of 12 representation layers of a BERT encoder, which layers encode the most word-level features? Third, the connection between word-level features embedded in BERT representations and the performance of downstream tasks such as named entity recognition (NER) and natural language inference (NLI). Intuitively, some tasks rely on word features more than other tasks. By analyzing word knowledge inside BERT after fine-tuning on each task, we can gain empirical evidence on how the task is solved by BERT.

We take two main approaches for analyzing BERT. First, BERT is based on Transformer ~\cite{transformer}, a multi-layer multi-head self-attention-network architecture, where the embedding of each character is calculated by neural attention~\cite{DBLP:journals/corr/BahdanauCB14} over all the characters in a sequence. Inspired by Clark et al.~\shortcite{attvis}, we look into the attention weight distribution of each head in each layer, in order to understand whether some attention head exhibit salient word-level patterns in the attention targets. Second, we take Chinese word segmentation as a probing task~\cite{probe1,probe2}, training a linear classifier based on the BERT representation at each Transformer layer, so that word-level information contained in the layer can be quantified through segmentation performance.

Results on two Chinese datasets with varying segmentation standards consistently show that word information is captured by BERT representation. There exist attention heads that focus on the start and end characters in each word, word unigrams and bigrams, as well as word boundary patterns. In addition, word information is captured mostly in the middle layers of Transformer, allowing light-weight probing layers to achieve segmentation F1 score around 90\% on both segmentation datasets. Finally, we find that different Chinese tasks require different levels of word information, with fine-tuning on tasks such as POS tagging and chunking significantly improving the probing task, while fine-tuning on tasks such as NLI significantly decreasing probing accuracies.

To our knowledge, we are the first to investigate  word structure knowledge in Chinese BERT.  Our code has been released at \url{https://github.com/ylwangy/BERT_zh_Analysis}.

\section{BERT}

BERT~\cite{bert} consists of multi-layer Transformer~\cite{transformer} blocks. Formally, given a sentence  $s = c_1, c_2, ..., c_{n}$, each $c_i$ is first transformed into input vector $e_i$ by summing up token embeddings \textit{$E_{c_i}$}, segment embeddings $\it{SE}_{c_i}$, which distinguish the sentence location, and position embeddings $\it{PE}_{c_i}$, which indicates character position:
\begin{equation}
e_i = E_{c_i} + \it{SE}_{c_i} + \it{PE}_{c_i} 
\label{eq:e}
\end{equation}

The vectors $e_1, ..., e_n \in \mathbb{R}^{n\times{d}}$ are taken as input to the first layer in a Transformer encoder, which consists of $L$ layers. Now for each layer, denote the input as $E$. $E$ is then transformed into vectors for queries $Q^m$, keys $K^m$, and values $V^m$ via linear mappings, $\{Q^m, K^m, V^m\} \in \mathbb{R}^{n\times{d_k}}$:
\begin{equation}
Q^m, K^m, V^m = EW_Q^m, EW_K^m, EW_V^m ,
\label{eq:transformer1}
\end{equation}
where $\{W_Q^m, W_K^m, W_V^m\} \in \mathbb{R}^{d\times{d_k}}$ are trainable parameters, $m\in[1,2,...,M]$ represent the $m$-th attention head. $M$ parallel attention functions are applied to produce $M$ hidden states $\{H^1, ..., H^M\}$:
\begin{equation}
\begin{array}{l}
{\alpha}^m = \it{softmax}(\frac{Q^{m}{K}^{m\top}}{\sqrt{d_k}}) \\
H^m = {\alpha}^mV^m
\end{array}
\label{eq:transformer2}
\end{equation}

$\alpha^m$ is the attention distribution for the $m$-th head and $\sqrt{d_k}$ is a scaling factor. Finally, multi-head hidden states are concatenated to obtain a hidden representation $\hat{h}_i$ of each character $c_i$:
\begin{equation}
\hat{h}_i = [H^1_i, ..., H^M_i]
\label{eq:h}
\end{equation}

$\hat{h}_i$ are then fed to a multi-layer perceptron for computing the final outputs $h_i$ for the layer. Feed-forward connections and layer normalization are also applied, the detail of which can be found in Vaswani et al.~\shortcite{transformer}. We denote the output of the $l$-th layer as $h^l_i$ ($l \in [1,2,...,L]$).

Given a corpus $\{\textit{$s_t = c_{1}, c_{2}, ..., c_{n_t}$}\}|_{t=1}^{T}$, the masked language model objective is to minimize the loss of predicting the randomly chosen masked character $c_{mask_j}$ ($j \in [1,2,...,n_t]$) by its representation $h_{mask_j}$ in the last layer $L$:
\begin{equation}
L_{\it{MLM}} = -\sum_{t=1}^{T}\sum_{j=1}^{n_t}{\log p(E_{c_{mask_j}}| h^L_{mask_j})} ,
\end{equation}
where $E$ is the token embedding table in Eq.~\ref{eq:e}.
%$p(E_{c_{mask_i}}| h^L_{mask_i})$ is calculated as a softmax probability distribution over the whole vocabulary $D$:
%\begin{equation}
%p(E_{c_{mask_i}}| h^L_{mask_i}) = \frac{\exp({{E_{c_{mask_i}}}^\top{h^L_{mask_i}})}}{\sum_{c_{k} \in D}\exp({{E_{c_{k}}}^\top{h^L_{mask_i}})}}
%\end{equation}

During pre-training, special tokens [CLS] and [SEP] are added to indicate the beginning of a sentences and the separation of two sentences, respectively. %The [CLS] is important for many single-sentence tasks, and the [SEP] is used for the next sentence prediction objective. 
%The final pre-trained model we choose is
We conduct our experiments on BERT-base-Chinese\footnote{\url{https://github.com/google-research/bert}}, which has 12 layers, 12 heads and 768 hidden size. 

\begin{figure}[t]
	\subfigure[Attention to specific characters.]{
		\begin{minipage}[t]{1\linewidth}
			\centering
			\includegraphics[scale=0.9]{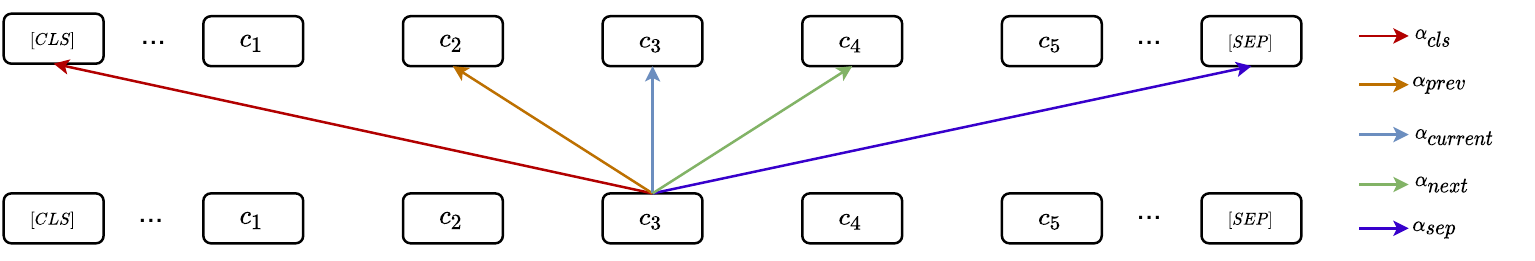}
			\label{figure:charatt}
		\end{minipage}%
	}
	\subfigure[Attention to word boundary characters.]{
		\begin{minipage}[t]{1\linewidth}
			\centering
			\includegraphics[scale=0.9]{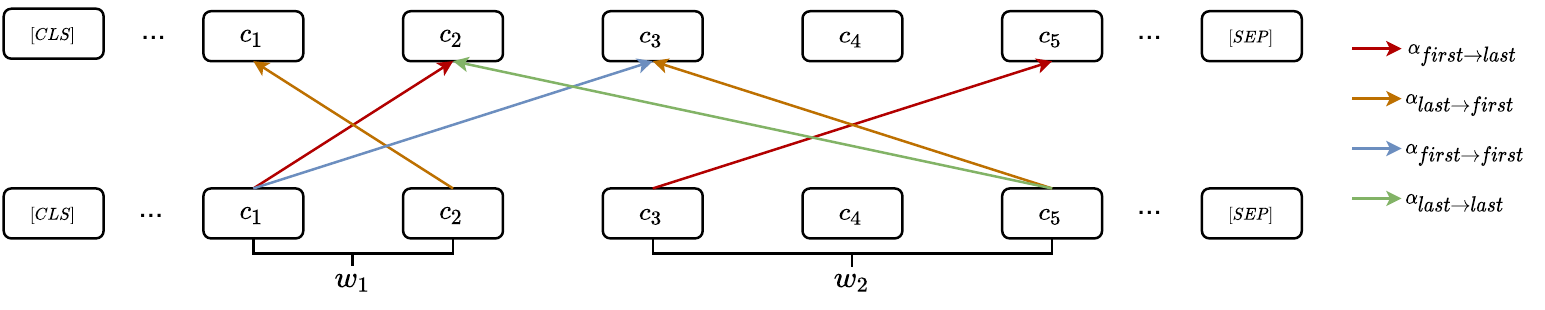}
			\label{figure:wordatt}
		\end{minipage}
	}
	\subfigure[Averaged attention to curruent and surrounding words.]{
		\begin{minipage}[t]{1\linewidth}
			\centering
			\includegraphics[scale=0.9]{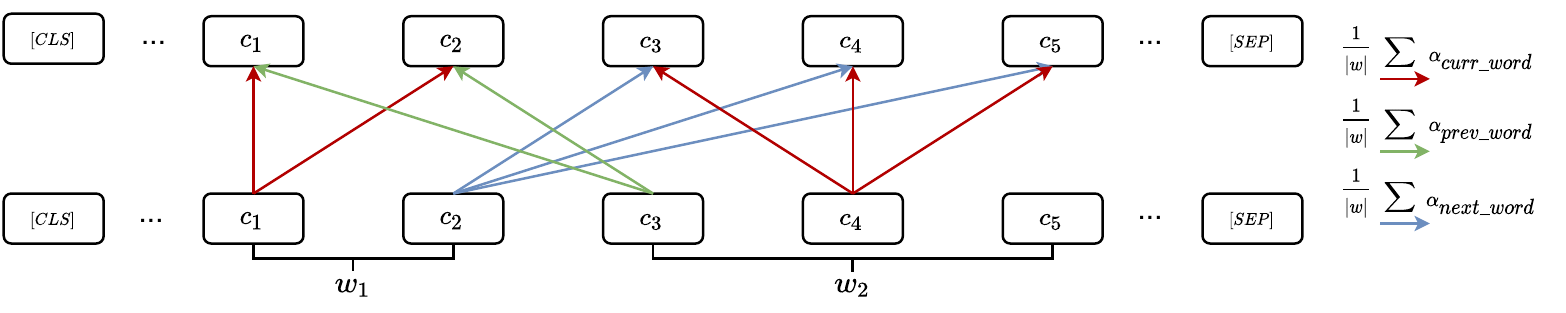}
			\label{figure:wordaccatt}
		\end{minipage}
	}
	\caption{Different attention weight distribution patterns.}
	\label{figure:att}
\end{figure}
\section{Attention Distribution Analysis}
We analyze the distribution of attention weight $\alpha$ in Eq.~\ref{eq:transformer2} across different layers and heads. Specifically, we compute the attention weight from characters to several specific characters as well as the attention distribution from characters to surrounding words. The structures are illustrated by  Figure~\ref{figure:att}.\footnote{We mainly show the difference of the attention patterns for character $c_i$, and omit the query, key and value transformations.}

For each character, we first investigate its attention weights to several specific characters, including the character itself, the previous character, the next character, and two special tokens [CLS] and [SEP], as shown in  Figure~\ref{figure:charatt}.
Formally, the attention from $c_i$ to $c_j$ in a certain head is defined as in Eq.~\ref{eq:transformer2}:

\begin{equation}
\alpha_{c_i \to c_j} = \frac{\it{exp}(Q_i{K}^{\top}_j/\sqrt{d_k})}{\sum_l{\it{exp}(Q_i{K}^{\top}_l/\sqrt{d_k})}} ,
\label{eq:speci}
\end{equation}
where $c_j$ can be the previous, current, or next character $c_{i-1}$, $c_i$, $c_{i+1}$, and the token [CLS] or [SEP], respectively.

We further investigate attention to characters at word boundary locations. Formally, for each $c_i$ that is the first character of some word $w_k$, we first consider the attention weights to the last characters of the current word and the first character of the next word:

\begin{equation}
\begin{array}{l}
\begin{aligned}
&\alpha_{c_i: first\to last} = \alpha_{c_i \to c_j}, s.t.\  w_k=\{c_i,...,c_j\} \\
&\alpha_{c_i: first\to first} = \alpha_{c_i \to c_j}, s.t.\  w_k=\{c_i,...\}, w_{k+1}=\{c_j,...\}
\end{aligned}
\end{array}
\label{eq:first}
\end{equation}

Similarly, for each $c_j$ that is the last character of some word $w_k$, we consider the attention weights to the first character of the current word and the last character of the previous word:
\begin{equation}
\begin{array}{l}
\begin{aligned}
&\alpha_{c_j: last\to first} = \alpha_{c_j \to c_i}, s.t.\  w_k=\{c_i,...,c_j\} \\
&\alpha_{c_j: last\to last} = \alpha_{c_j \to c_i}, s.t.\  w_{k-1}=\{...,c_i\}, w_{k}=\{...,c_j\}
\end{aligned}
\end{array}
\label{eq:last}
\end{equation}

Figure~\ref{figure:wordatt} shows one example, where $w_1$ and $w_2$ are two adjacent words,  $c_1$, $c_2$ are the first and last character of the word $w_1$, and $c_3$, $c_5$ are the first and last character of the word $w_2$, respectively. For this example, we consider the attentions  $\alpha_{c_1 \to c_2}$,  $\alpha_{c_3 \to c_5}$, $\alpha_{c_1 \to c_3}$ (Eq.~\ref{eq:first}), and $\alpha_{c_2 \to c_1}$,  $\alpha_{c_5 \to c_3}$,  $\alpha_{c_5 \to c_2}$ (Eq.~\ref{eq:last}).

\begin{table}[!t]
	\centering
	\small
	\subfigure[Results on CTB9.]{
		\begin{minipage}{1\linewidth}
			\centering
			\small
			%\resizebox{\textwidth}{20mm}{
			\begin{tabular}{c|c|c|c|c|c|c|c|c|c}
				\hline 
				\multirow{2}{*}{\bf{Layers}}&  \multicolumn{5}{c|}{\bf{Specific Characters}} &  \multicolumn{4}{c}{\bf{First\&Last Character of Words}}  \\ 
				\cline{2-10}
				& Curr &Next & Prev &CLS & SEP  &  F$\to$L &L$\to$F & F$\to$ F$_1$ &L$\to$L$_{\textnormal{-}1}$\\
				\hline
				\#1 &  9.0  & \textbf{62.0}(5)& 35.7 &35.3  &  12.6  &  15.7 & 13.7&  7.2&7.7 \\
				\hline
				\#2 &  7.6  & 60.8&62.7 &  60.8 &  12.1  &  26.7 & 28.7&  19.4 & 21.9 \\
				\hline
				\#3 &  38.7 & 31.5&61.5 & \textbf{65.5}(4)  &  25.2  & \textbf{64.7}(10) &\textbf{59.7}(11)&  7.7 & 9.1\\
				\hline
				\#4 &  44.5  & 61.5&38.0 & 21.0  &  61.4  & 38.0 &23.3& 16.0 & 23.6 \\
				\hline
				\#5 &  7.9  & 13.4&17.7 & 25.8 &  80.1  &   19.7 &39.7& 14.2 &12.7\\
				\hline
				\#6 &  12.6  & 36.7&33.5 & 9.8  &  57.3 &  25.8& 56.4&  20.9 &18.3\\
				\hline
				\#7 &  5.4  & 37.4&\textbf{91.9}(4) & 4.1  &  78.1  & 60.9 & 38.2&  \textbf{33.3}(1) & 25.2\\
				\hline
				\#8 &  6.7 &52.8& 34.4 & 8.7  & 90.4  & 35.3 & 33.8& 24.1 & \textbf{26.8}(3)\\
				\hline
				\#9 &  9.9  & 18.7&12.1 & 13.8  & 86.4  &  35.1 & 55.3& 9.3&13.1\\
				\hline
				\#10 &  18.5 & 9.2& 9.4 & 5.6  & 88.7  &  42.2& 37.6& 9.0 &10.5\\
				\hline
				\#11 &  12.8  & 32.0&15.8 & 11.0  &  88.1  & 19.5 & 24.2& 5.9 &8.32\\
				\hline
				\#12 &  \textbf{55.1}(2)  & 7.1& 3.3 &1.6  &  \textbf{90.7}(6)  & 8.0& 8.7& 7.3 &6.6\\
				\hline
			\end{tabular}
			%}
		\end{minipage}%
	}
	\subfigure[Results on PKU.]{
		\begin{minipage}{1\linewidth}
			\centering
			\small
			%\resizebox{\textwidth}{20mm}{
			\begin{tabular}{c|c|c|c|c|c|c|c|c|c}
				\hline 
				\multirow{2}{*}{\bf{Layers}}&  \multicolumn{5}{c|}{\bf{Specific Characters}} &  \multicolumn{4}{c}{\bf{First\&Last Character of Words}}  \\ 
				\cline{2-10}
				& Curr &Next & Prev &CLS & SEP  &  F$\to$L &L$\to$F & F$\to$ F$_1$ &L$\to$L$_{\textnormal{-}1}$\\
				\hline
				\#1 &  8.6  & 58.6& 34.1 &34.7  &  11.8  &  13.7 & 10.9&  6.3&6.2 \\
				\hline
				\#2 &  6.7  & 58.9&61.1 & 62.0&  11.8  &  23.8 & 22.3&  17.0 & 22.5 \\
				\hline
				\#3 &  36.0 & 34.9&62.8 & \textbf{68.7}(4)  &  23.1 & \textbf{58.4}(10) &50.3& 5.3 & 9.2\\
				\hline
				\#4 &  43.4  & 64.1&39.4 & 20.0 & 56.5 & 36.7 &17.4& 14.9 & 21.6 \\
				\hline
				\#5 &  9.7  & 14.3&17.8 &26.3 &  74.9  &   16.4 &33.1& 12.4 &10.2\\
				\hline
				\#6 &  13.0  & 41.5&33.8 & 10.1  &  56.4 &  24.8&\textbf{50.5}(5)&  18.7 &18.5\\
				\hline
				\#7 & 5.8 & 38.1&\textbf{91.7}(4)& 4.2  & 82.3  & 55.1 & 34.9&  \textbf{29.5}(1) & 22.3\\
				\hline
				\#8 &  6.5&\textbf{60.1}(5)&38.2 &7.0  & \textbf{91.0}(11)  & 36.7 & 31.4&19.3 & \textbf{24.7}(3)\\
				\hline
				\#9 & 11.0&22.7&13.6 &12.5  & 83.7 &  28.3 &51.1&7.4&11.0\\
				\hline
				\#10 & 16.4 & 10.7&11.8 & 6.4  &88.7   &  39.5& 34.5& 7.4 &9.4\\
				\hline
				\#11 &  16.8  & 36.4&17.2 & 6.8  & 90.3 & 18.8 & 22.4& 4.3 &7.2\\
				\hline
				\#12 & \textbf{60.3}(2)  & 7.5& 2.7 &2.2  &  87.3 & 5.6& 6.2& 4.7 &4.9\\
				\hline
			\end{tabular}
			%}
		\end{minipage}%
	}
	\caption{ Character-to-character attention distribution. The numbers $j$ in the parentheses denotes the $j$-th head. F, L, F$_1$, and L$_{\textnormal{-}1}$ denote first character of current word, last character of current word, first character of next word, and last character of previous word, respectively.}
	\label{table:attresult}
\end{table}

In addition to character-to-character attention weights, we consider character-to-word attention by taking the average of attention weights to characters that belong to specific words. Formally, we define the attentions from a character $c_j$ in word $w_k$ to it previous, current and next words as follows:
\begin{equation}
\begin{array}{l}
\begin{aligned}
&\alpha_{c_j: {prev\_word}} =\frac{1}{|w_{k-1}|} \sum_{c_i \in w_{k-1}}\alpha_{c_j \to c_i}\\
&\alpha_{c_j: {curr\_word}} =\frac{1}{|w_{k}|}  \sum_{c_i \in w_{k}}\alpha_{c_j \to c_i}\\
&\alpha_{c_j: {next\_word}} =\frac{1}{|w_{k+1}|}  \sum_{c_i \in w_{k+1}}\alpha_{c_j \to c_i} ,
\end{aligned}
\end{array}
\label{eq:att2words}
\end{equation}
where $|w|$ denotes the length of word $w$.

Figure~\ref{figure:wordaccatt} shows one example, where the $\alpha_{curr\_word}$ for $c_1$ is the average of  $\alpha_{c_1 \to c_1}$ and  $\alpha_{c_1 \to c_2}$, because $c_1$ belongs to the words $w_1$ composed of characters $c_1$ and $c_2$. $\alpha_{next\_word}$ for $c_1$ is the average of  $\alpha_{c_1 \to c_3}$ ,  $\alpha_{c_1 \to c_4}$ and  $\alpha_{c_1 \to c_5}$.

\subsection{Datasets}
There exist different word segmentation criteria for the same sentence, which mainly differ by the segmentation granularity. We select two corpora with golden word segmentation labels for computing attention distribution: Chinese Treebank~\cite{xue2005penn} 9.0 (CTB9) and PKU~\cite{sighan}, averaging the attention weights across all the characters for each attention heads.

%and PKU from SIGHAN bakeoff~\cite{sighan}
\subsection{Results}
%\textbf{Character-to-Character Attention.}
%\paragraph{\bf Character-to-Character Attention.}
\noindent \textbf{Character-to-Character Attention.} Table~\ref{table:attresult} show the largest values among all the 12 attention heads in each layer of BERT. We use head $i$-$j$ to denote the $j$-th attention head in the $i$-th layer. The average sentence lengths of the CTB9 and PKU datasets are 26.3 and 35.8, respectively. As a result, random baseline for the two datasets are 3.8\% (1/26.3) and 2.7\% (1/35.8), respectively.

%For specific characters, we find that few of the attention heads focus on the current character. However, the head 12-2 puts 55.1\% attention to itself on CTB9. The values increase when considering the next and previous characters. For example,
%for the neighboring next and previous tokens increased, 
%the head 7-4 focuses on the next character with more than 90\% attention weight. For the [CLS] and [SEP] tokens, values for [SEP] are much higher.  These findings are consistent with Clark et al.~\shortcite{attvis}, showing the different character-level preference from different heads. The distribution results of CTB9 and PKU are similar in general, although the heads can be different. For example, the best head attending to the next character is at layer 1 in CTB9 but layer 8 in PKU. 

\begin{figure}[!t]
	\subfigure[]{
	\begin{minipage}[t]{0.5\textwidth}
		\centering
	\includegraphics[scale=0.186,trim= 155 0 155 0]{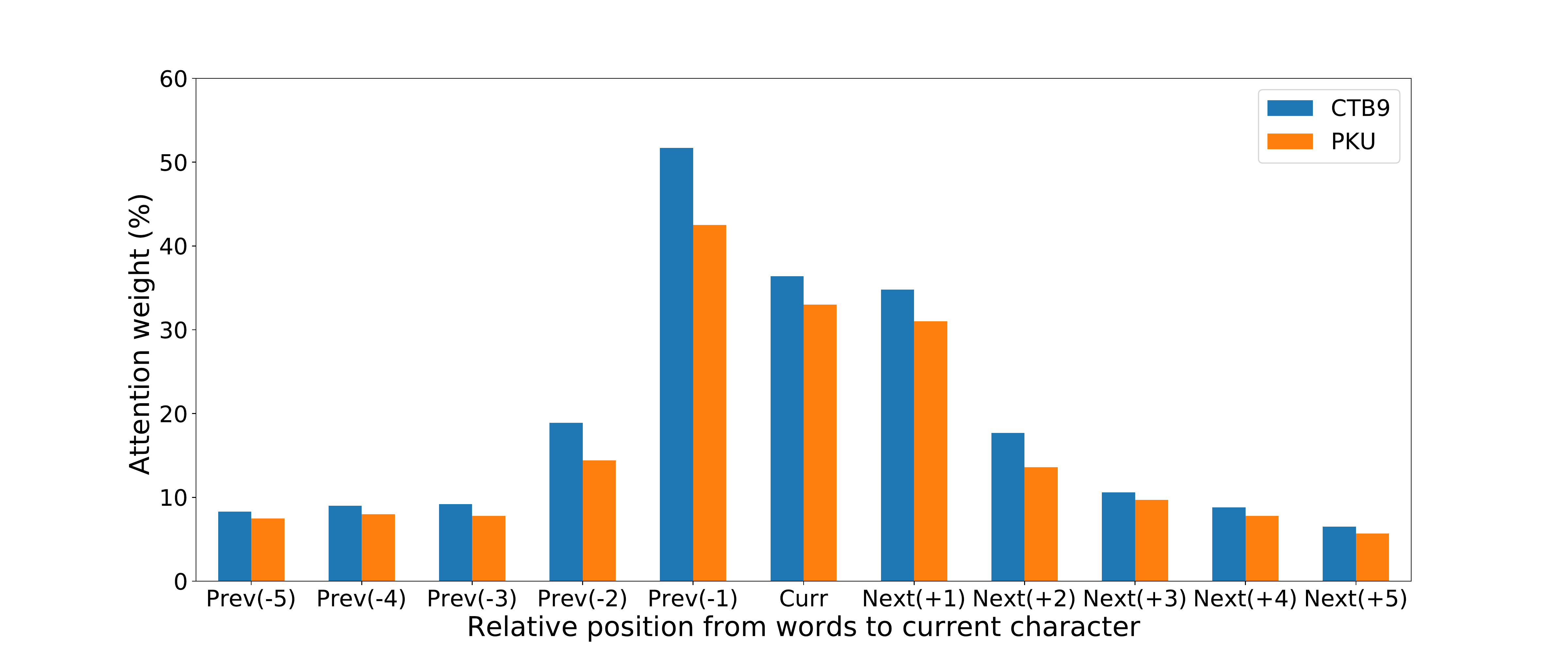}
	\label{figure:char2wordatt}
\end{minipage}}
	\subfigure[]{
	\begin{minipage}[t]{0.5\textwidth}
		\centering
	\includegraphics[scale=0.186,trim= 155 0 155 0]{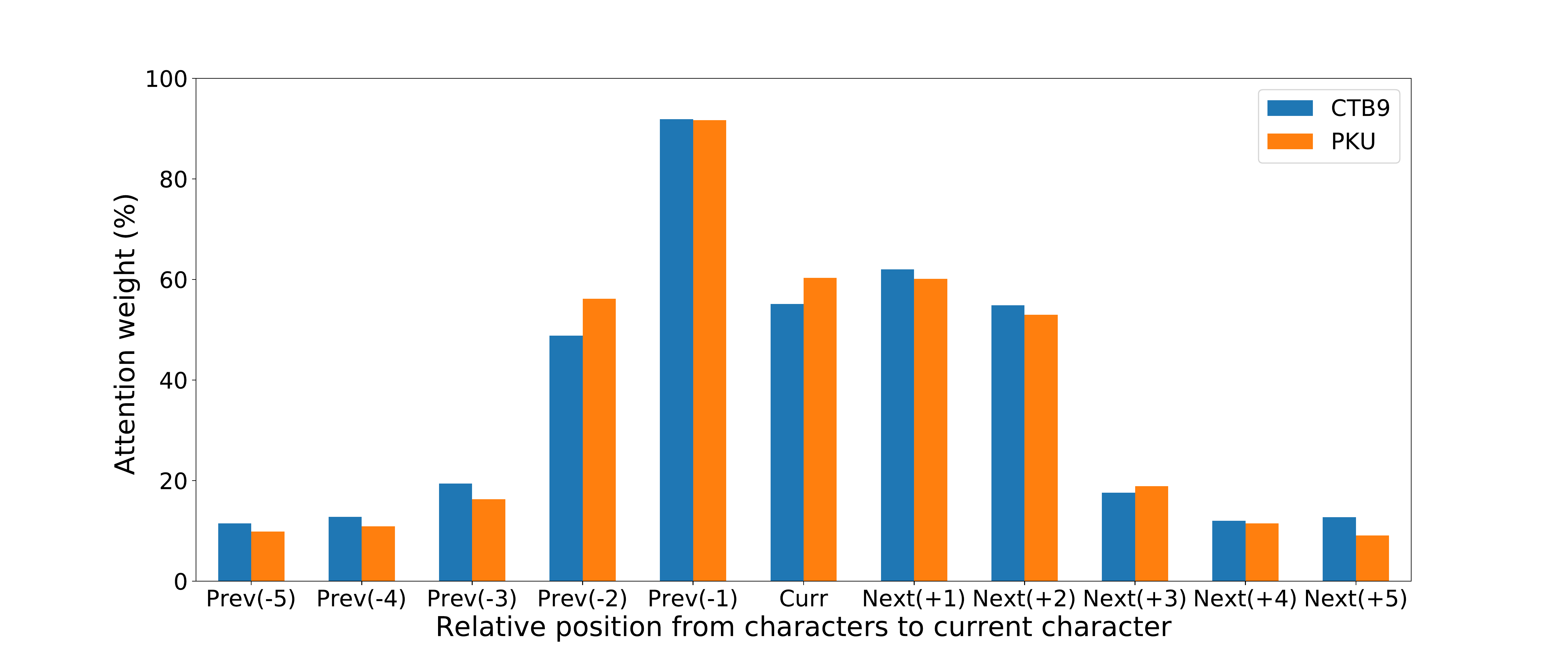}
	\label{figure:char2charatt}
\end{minipage}}
\caption{Character-to-word and character-to-characters attention distribution with different positions.}
\end{figure}
%For internal word characters ($\alpha_{first\to last}$ and $\alpha_{last\to first}$), two heads from layer 3 give the best results, taking 64.7\% and 58.4\% attention weight $\alpha_{first \to last}$ in CTB9 and PKU datasets, respectively. For external word characters ($\alpha_{first\to first}$ and $\alpha_{last\to last}$), the best values (33.3\% and 29.5\%) become much lower. The reason can be that the external word context are much more sparse compared with internal word context. Nevertheless, considering the average sentence length in CTB9 and PKU are 26.3 and 35.8, the results are still significantly better than the random uniform distribution results of 3.8\% and 2.7\%, respectively.

%The average sentence lengths of the CTB9 and PKU datasets are 26.3 and 35.8, respectively. As a result, random baeline for the two datasets are 3.8\% and 2.7\%, respectively.

First, the Specific Characters columns of the two tables show weights calculated by Eq.~\ref{eq:speci}. There are specific heads with strong attention weights to both the current character and its neighboring characters. For both datasets, attention to the previous character can reach 92\%, while to the next character is 60\%. This shows that the previous character plays a very important role in certain BERT representation layers. Attention weights to [SEP] and [CLS] are also significantly stronger than the random baseline, with those to the [SEP] node being above 90\%. This shows that BERT is highly sensitive to sentence boundaries. Finally, results are quite consistent between CTB9 and PKU.

The First\&Last Characters of Words columns of the two tables show attention weights calculated by Eqs.~\ref{eq:first} and~\ref{eq:last}. First, attention between the first and last characters of the same word can reach 50\% to 60\%, which shows that word information is captured by BERT representation. In addition, attention between consecutive words can reach 20\% to 30\%, which shows that word n-gram information is also learned. Word internal attention is higher than cross-word attention, which can be because word n-grams are more sparse and play less role in BERT.

Finally, the strongest weights mostly occur in the middle layers (3$\sim$8), which suggests that word information from BERT concentrates in those layers. Our probing task in Section 4 confirms this.

Our findings are in line with those by Clark et al.~\shortcite{attvis}, who observe that different heads in English BERT models can put strong attention weights on different dependency relations. In addition, Jawahar  et al.~\shortcite{analysis-bert-1} shows that syntax in BERT are in relatively lower layers while semantics are in higher layers. Word information can be regarded as a lexical level syntax feature, and thus our finding is similar.

%In spite of that, considering the average length of sentences is 18.6, the above results are still much better than the uniform distribution of  5.37\% (1/18.6), indicating the model can still learn the word information to some extend.

%\textbf{Character-to-Word Attention.}
%\paragraph{\bf Character-to-Word Attention.}
%Figure~\ref{figure:char2wordatt} shows the result of best-performing heads from character to the surrounding words according to Eq.~\ref{eq:att2words}. In addition to the current, previous and next words, we also show the results for other words  within a window of size 5. Some heads put more than 50\% of the attention weight to the characters in previous word. Interestingly, attention weight to the previous word is larger than those to the current and next word. For words far from the current character, the attention weight decreases, indicating that the words far away from the current character contribute less to its contextualized representation compared with the neighboring words. The best attention weights on CTB9 are  higher than PKU. The reason can be that the average sentence and word lengths of CTB9 are less than that of PKU (26.3\&1.56 \textit{v.s} 35.8\&1.64), thus making the attention distribution more concentrated.

\noindent \textbf{Character-to-Word Attention.} Figure~\ref{figure:char2wordatt} shows the results of the best-performing heads calculated using Eq~\ref{eq:att2words}. In addition to the current, previous and next words, we also show the results for other words within a window size of 5. We can find that attention to the previous word is stronger compared with that to the current word and the next word. Attention weights decrease when the target is increasingly far from the current characters, which shows that neighboring context is more important in BERT representation concerning words. This is consistent with character-to-character trends (Figure~\ref{figure:char2charatt}). Values for CTB9 are relatively larger than those for PKU mainly because the sentences are shorter, but the main observations are consistent.

\begin{CJK*}{UTF8}{gbsn}	
	\begin{figure}[t]
		\subfigure[Head 12-2]{
			\begin{minipage}[t]{0.18\linewidth}
				\centering
				\includegraphics[scale=0.3,trim= 80 60 60 0]{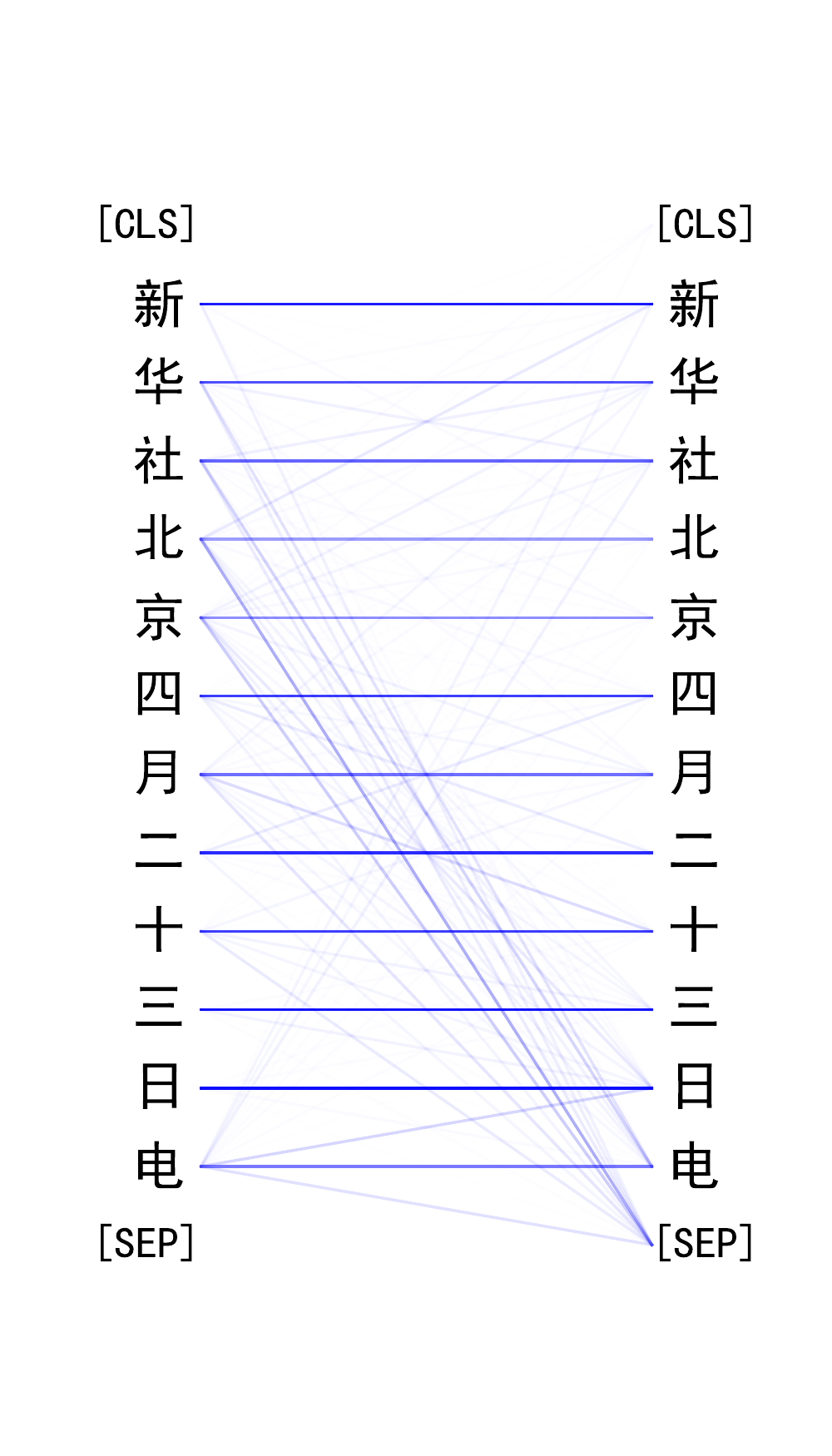}
				\label{figure:current}
			\end{minipage}%
		}
		\subfigure[Head 1-5]{
			\begin{minipage}[t]{0.18\linewidth}
				\centering
				\includegraphics[scale=0.3,trim= 80 60 60 0]{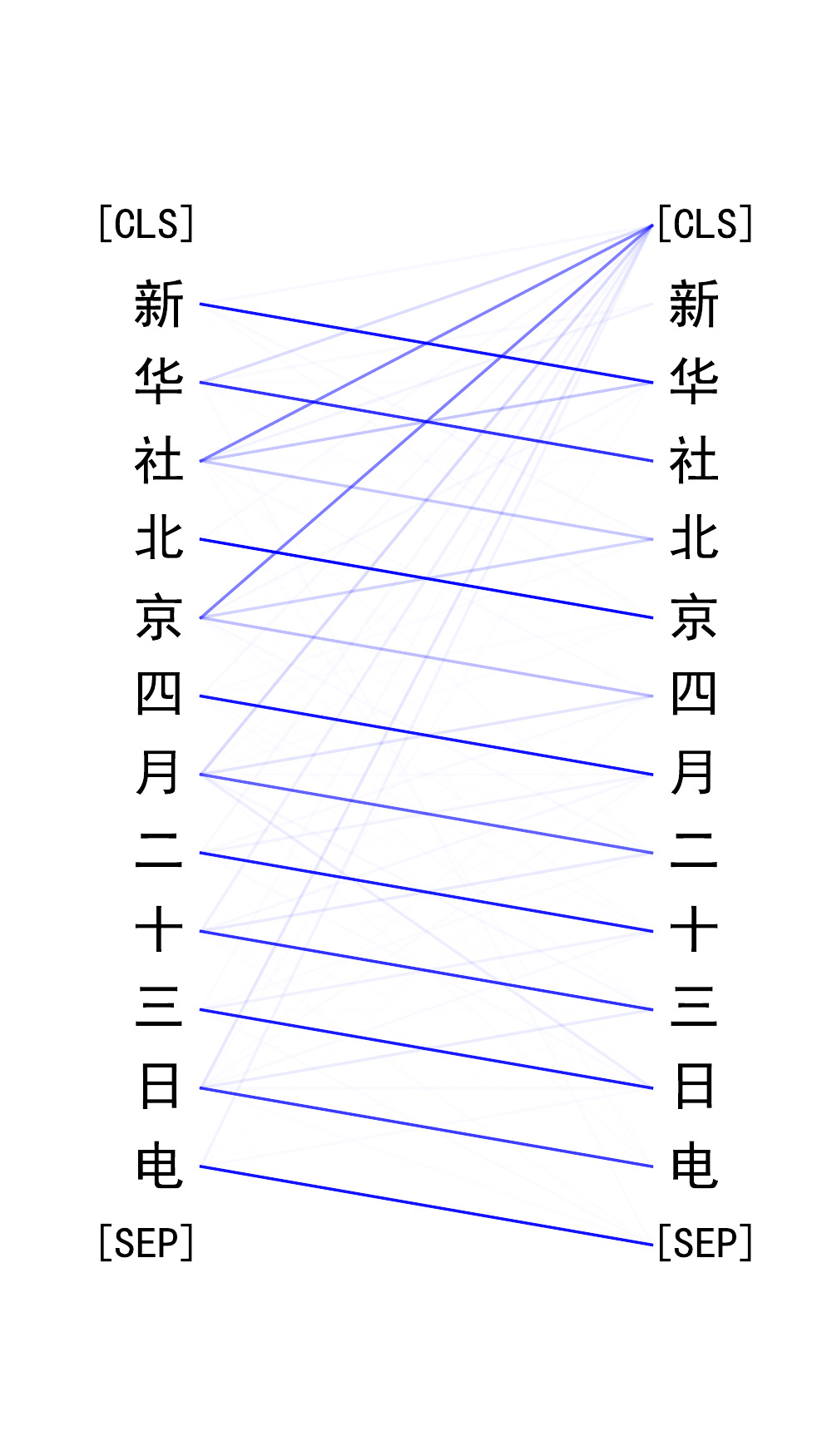}
				\label{figure:next}
			\end{minipage}
		}
		\subfigure[Head 7-4]{
			\begin{minipage}[t]{0.18\linewidth}
				\centering
				\includegraphics[scale=0.3,trim= 80 60 60 0]{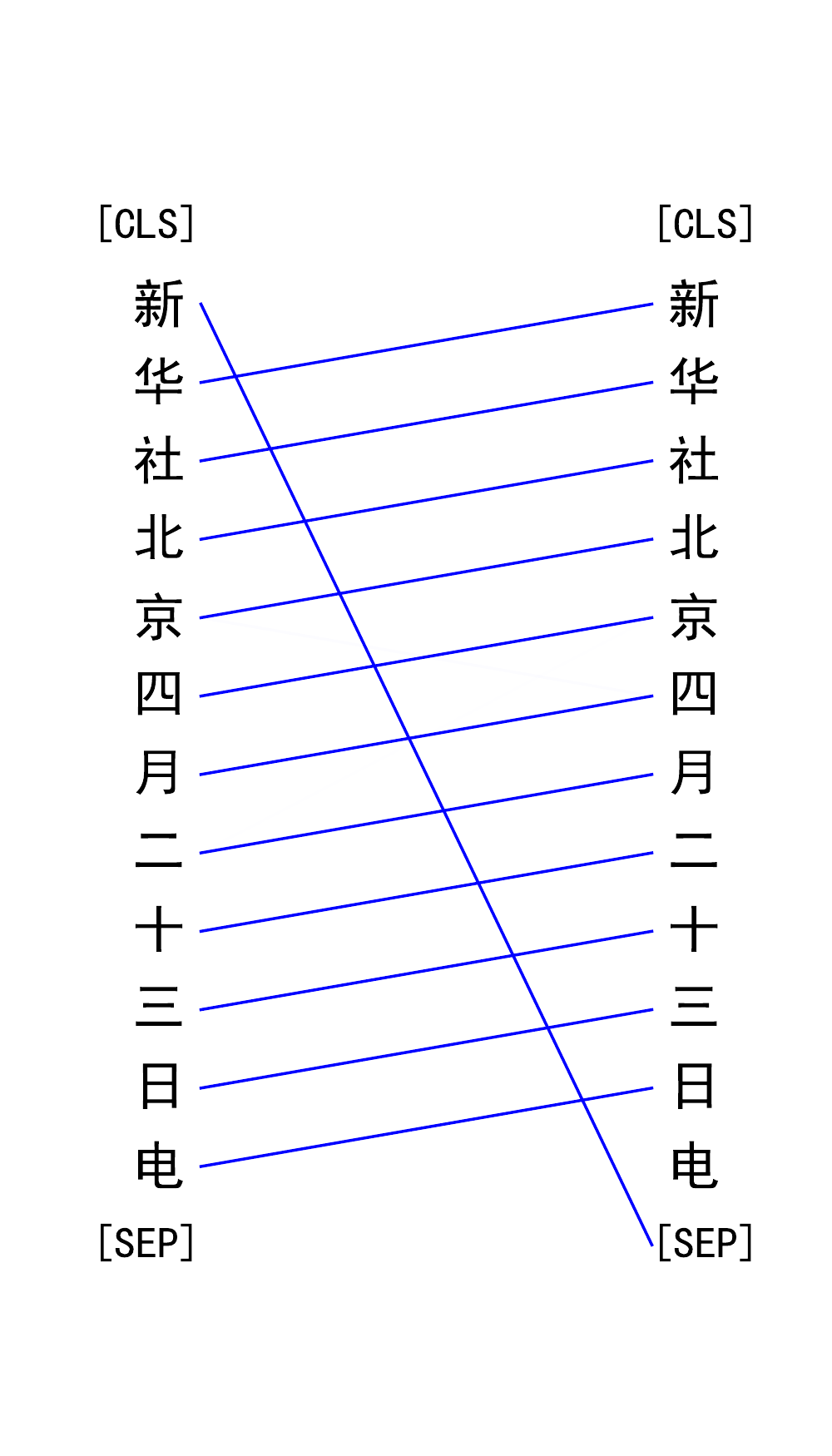}
				\label{figure:prev}
			\end{minipage}
		}
		\subfigure[Head 3-4]{
			\begin{minipage}[t]{0.18\linewidth}
				\centering
				\includegraphics[scale=0.3,trim= 80 60 60 0]{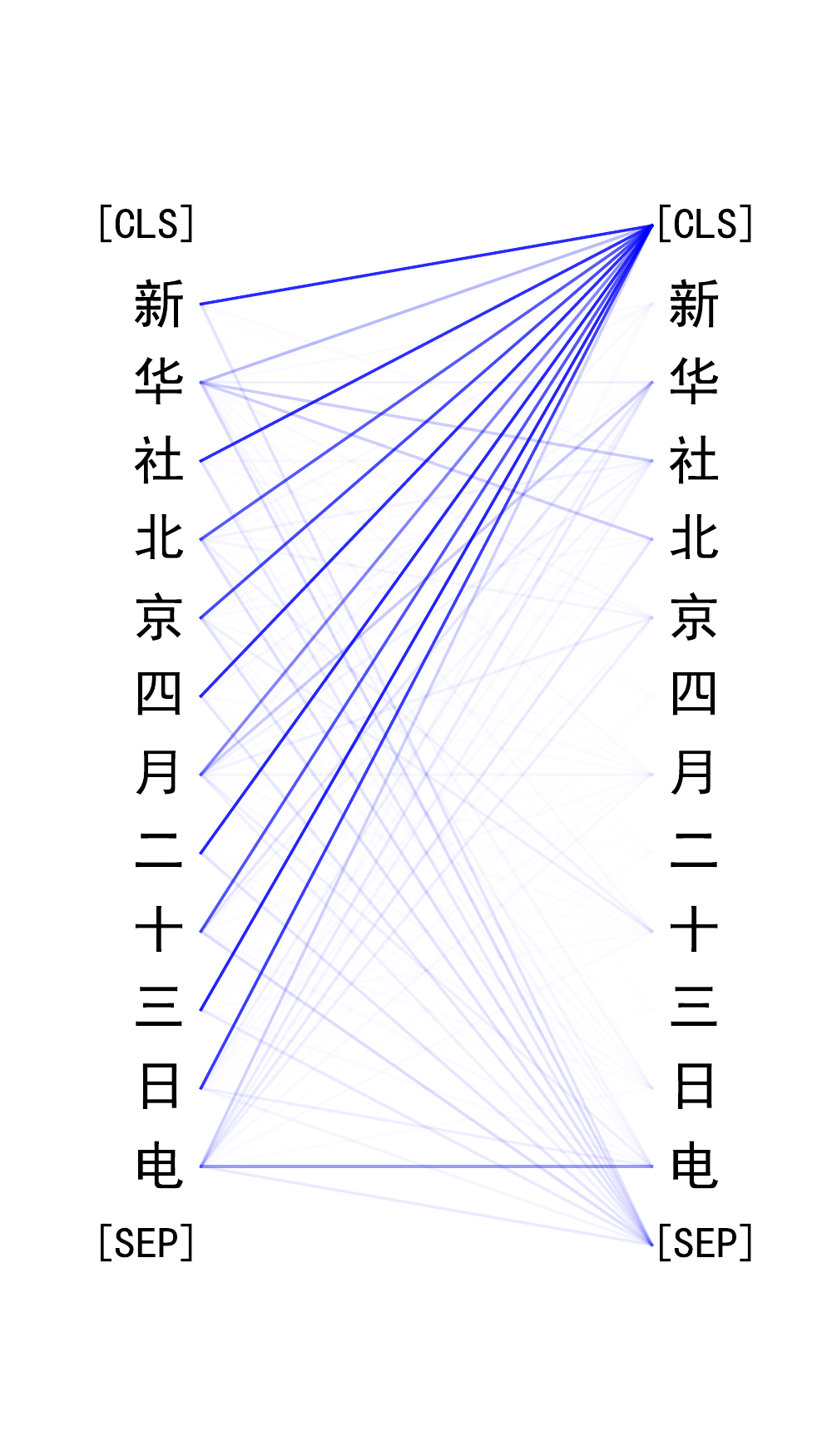}
				\label{figure:cls}
			\end{minipage}
		}
		\subfigure[Head 12-6]{
			\begin{minipage}[t]{0.18\linewidth}
				\centering
				\includegraphics[scale=0.3,trim= 80 60 60 0]{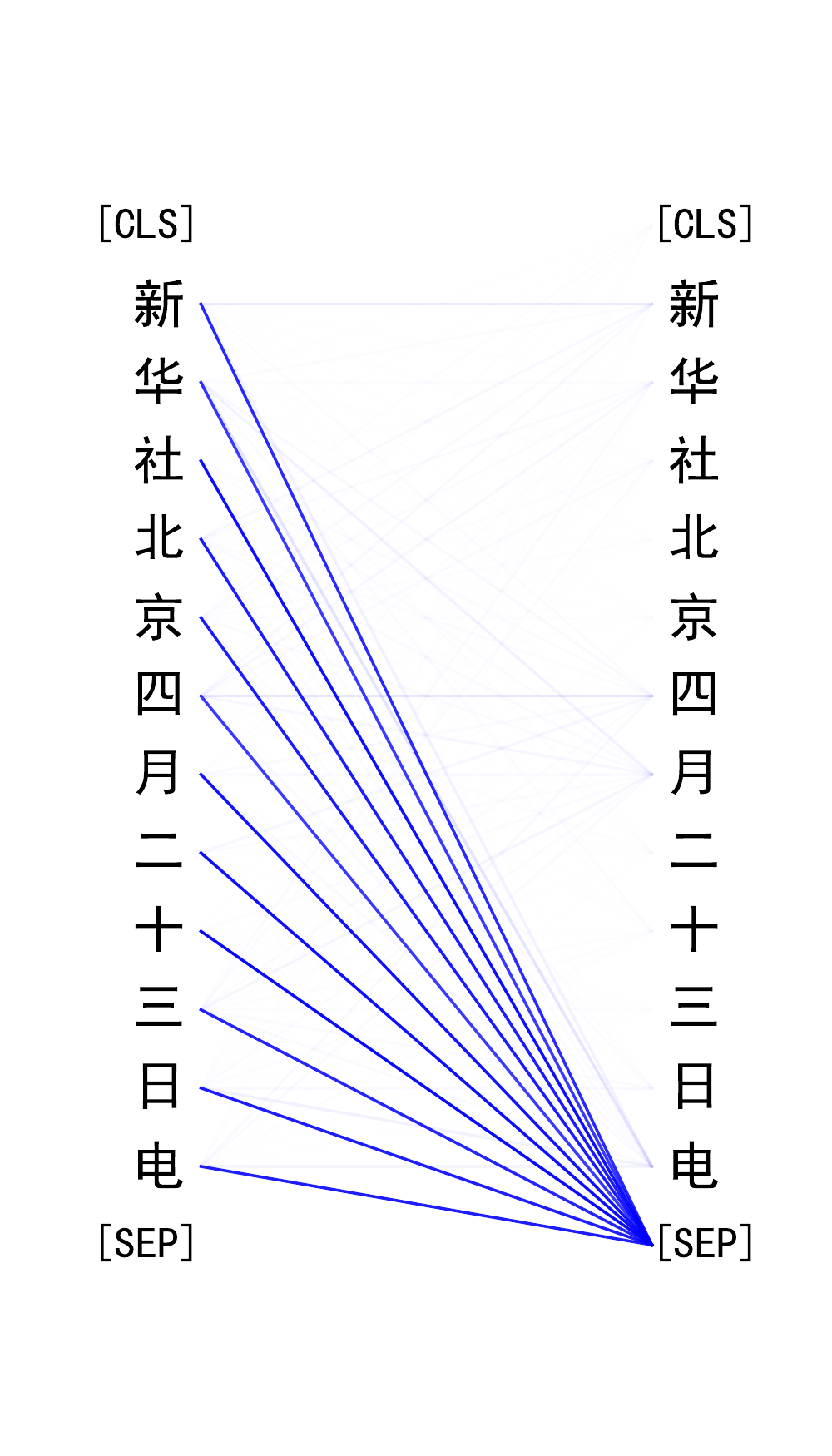}
				\label{figure:sep}
			\end{minipage}
		}%,trim= 80 60 60 0
		
		%\vspace{0.1cm}
		\subfigure[Head 3-10]{
			\begin{minipage}[t]{0.18\linewidth}
				\centering
				\includegraphics[scale=0.3,trim= 80 60 60 0]{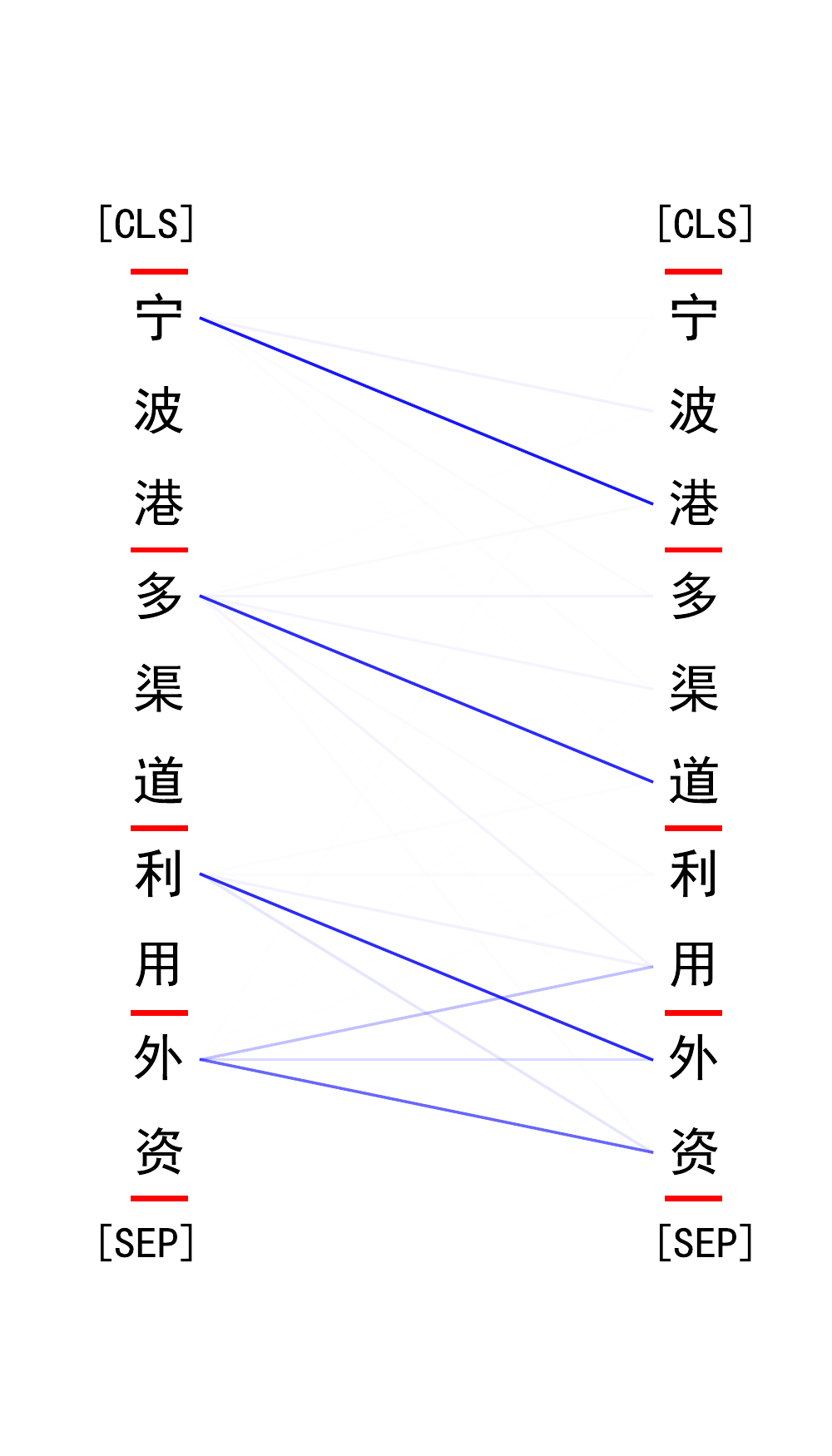}
				\label{figure:visattf} 
			\end{minipage}%
		}
		\subfigure[Head 3-11]{
			\begin{minipage}[t]{0.18\linewidth}
				\centering
				\includegraphics[scale=0.3,trim= 80 60 60 0]{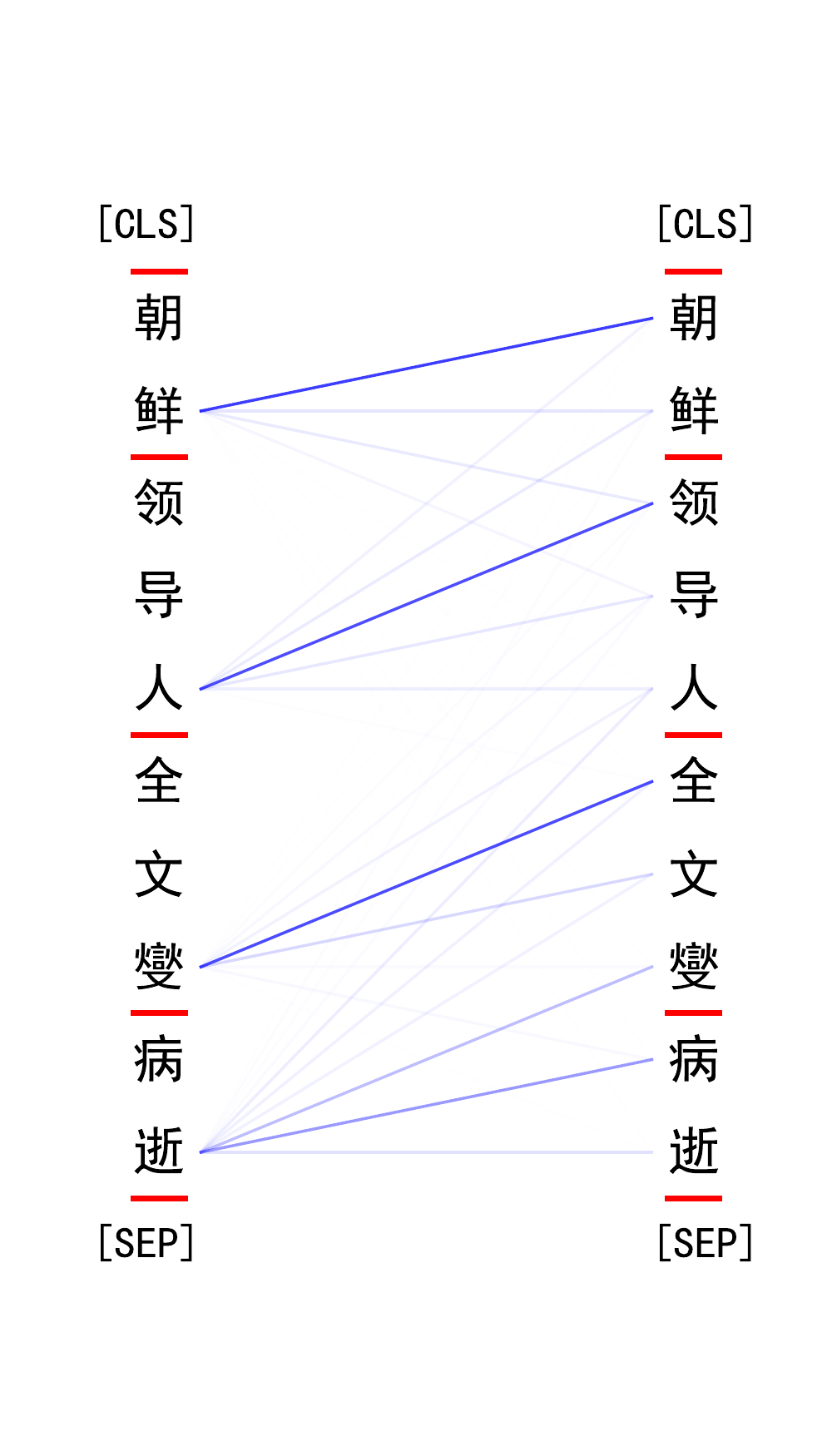}
				\label{figure:visattg}
			\end{minipage}%
		}
		\subfigure[Head 7-1]{
			\begin{minipage}[t]{0.18\linewidth}
				\centering
				\includegraphics[scale=0.3,trim= 80 60 60 0]{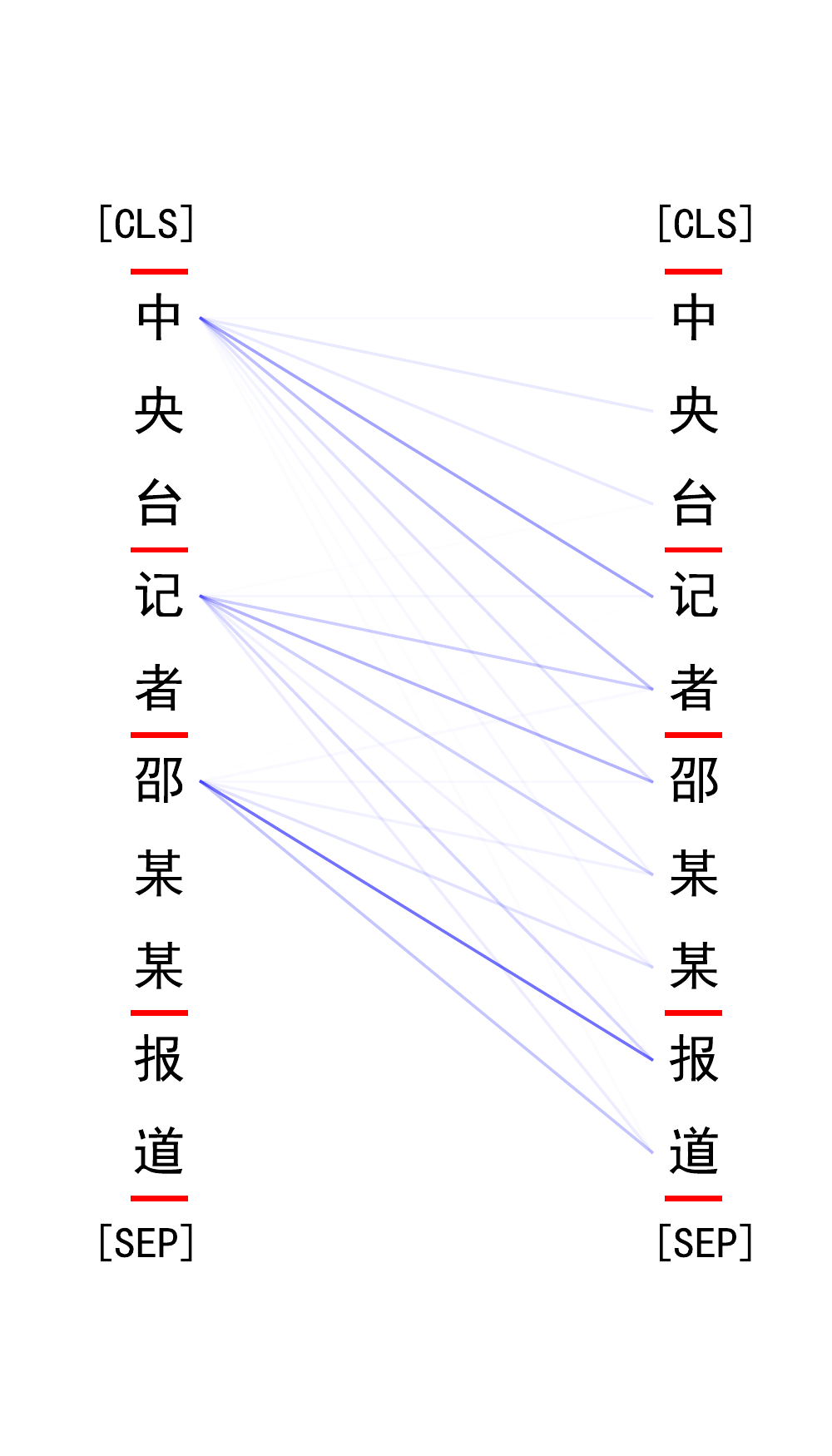}
				\label{figure:visatth}
			\end{minipage}
		}
		\subfigure[Head 8-3]{
			\begin{minipage}[t]{0.18\linewidth}
				\centering
				\includegraphics[scale=0.3,trim= 80 60 60 0]{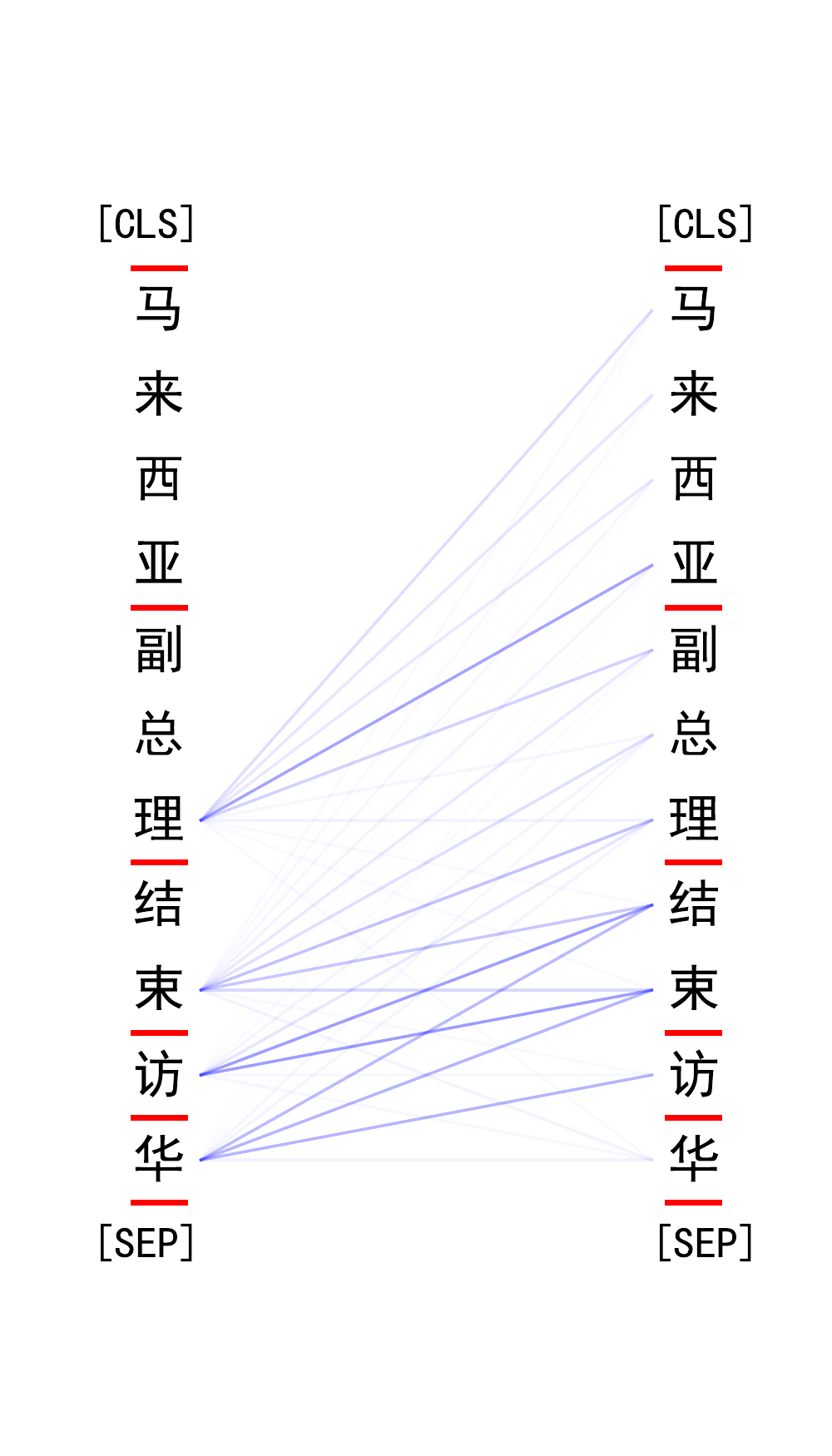}
				\label{figure:visatti}
			\end{minipage}%
		}
		\subfigure[Head 6-5]{
			\begin{minipage}[t]{0.2\linewidth}
				\centering
				\includegraphics[scale=0.3,trim= 80 60 60 0]{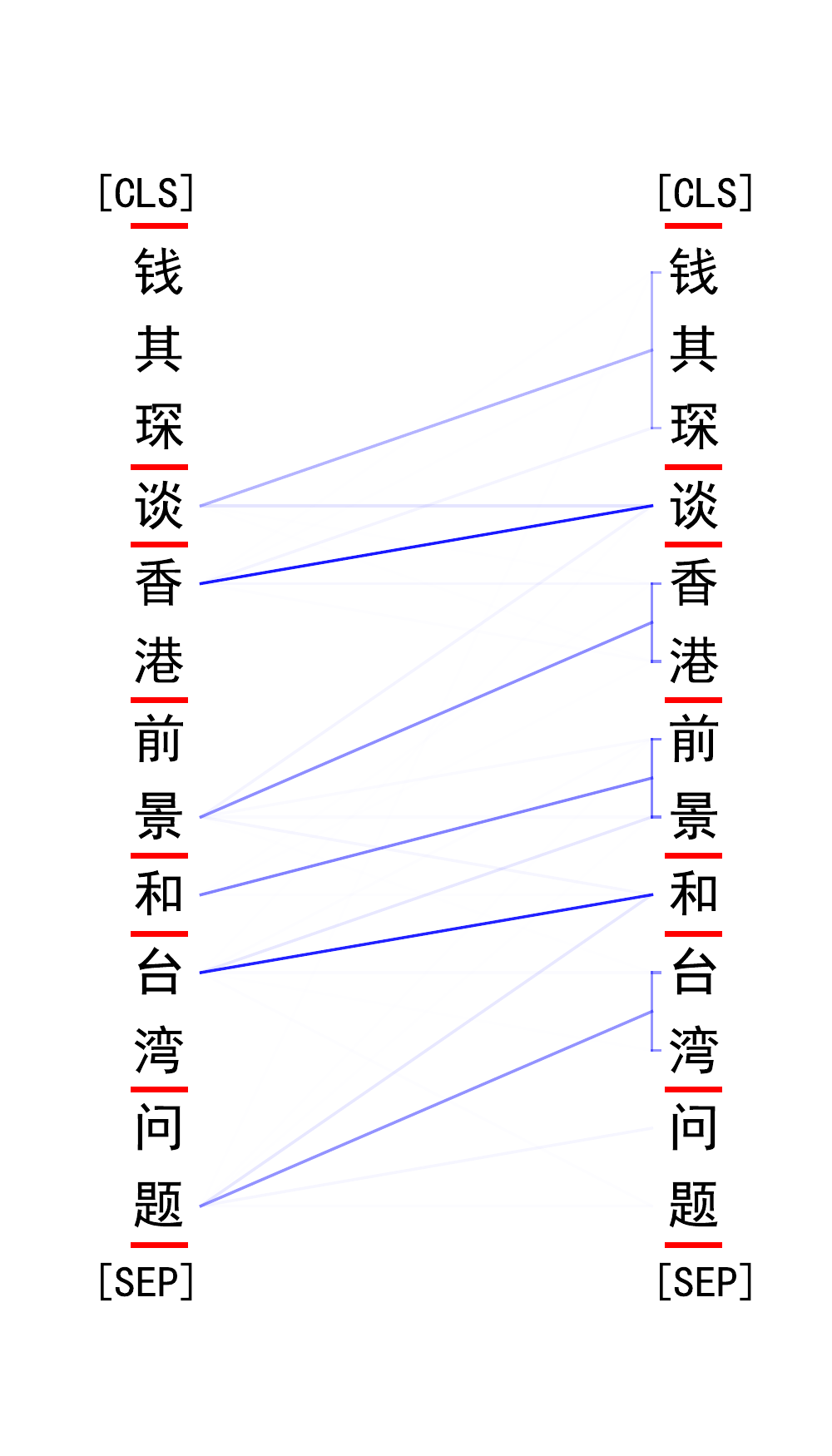}
				\label{figure:visattj}
			\end{minipage}%
		} %,trim= 70 60 60 0
		\caption{Examples of attention distribution visualization for sentences on CTB9. Head $i$-$j$ denotes the $j$-th attention head in the $i$-th layer.
			%新华社(Xinhua Agency)北京(Beijing)四月(April)二十三日(23th)电(news),宁波港(Ningbo Port)多渠道(multi-channel)利用(use)外资(overseas investment),朝鲜(North Korea) 领导人(leader) 全文燮(Wenxie Quan) 病逝(pass away),中央台(CCTV)记者(reporter)邵某某(Shao)报道(report),马来西亚(Malaysia)副总理(vice-premier)结束(end)访(visit)华(China),钱其琛(Qichen Qian)谈(talk)香港(Hong Kong)前景(prospect)和(and)台湾(Taiwan)问题(issue). 
			The darkness of blue lines reflect the values of attentions. The red lines denotes the word boundary.}
		\label{figure:visatt}
	\end{figure}
\end{CJK*}

%the word-level attentions, the best heads takes averaged 36.4\%, 34.8\% and 51.8\% attentions to characters in the current, next and previous words, respectively. But we find this mainly due to the fact these heads put most attentions to the current, next and previous characters. For example, the second head in layer 12 put 55.1\% attentions to the current token, thus we can not claim this head focus on certain word information. However, we still find some heads put attentions to words rather than specific common tokens. For example, for the fifth head in layer 6, it takes averaged 35.4\% attentions to the characters from the previous word, higher than that to the previous character. 

\subsection{Case Study}
\begin{CJK*}{UTF8}{gbsn}

We visualize the best-performing heads in Table~\ref{table:attresult} using a few sentences on CTB9. The top row of Figure~\ref{figure:visatt} shows the sentence ``新华社(Xinhua Agency)北京(Beijing)四月(April)二十三日(23th)电(news)". These heads consistently attend to one of the specific characters. For example, the head 1-5 attends to the next characters, with the last character ``电(news)" attending to the [SEP] token. The head 7-4 attends to the previous character, with the first character ``新(Novel)" attending to the [SEP] token as well.

Figures 3(f) to 3(i) show more examples that the heads attend to word boundary characters. For head 3-11, most of the last characters tend to pay attention to the first character of the same words. As for head 8-3, the last characters  attend to the last characters of the previous words. There are exceptions to the general trends above. Take head 3-10 for example, in the sentence ``宁波港(Ningbo Port)多渠道(multi-channel)利用(use)外资(overseas investment)", the character ``利(benefit)" puts the most attention on the character ``外(outside)", which is out of the corresponding word ``利用(use)".

Figure~\ref{figure:visattj} shows the attention distribution in the sentence ``钱其琛(Qichen Qian)谈(talk)香港(Hong Kong)前景(prospect)和(and)台湾(Taiwan)问题(issue)" for head 6-5. Each character puts the most attention on  characters in the previous word. For example, the character ``谈(talk)" focuses mainly on characters in the  word ``钱其琛(Qichen Qian)", and the character ``题(question)" puts most attention to characters in the word ``台湾(Taiwan)". This indicates that the head 6-5 takes most of the information from the previous words to generate contextualized character representation.
	
\end{CJK*}

\section{Probing Task}

\begin{figure}[!t]
	\centering
	\includegraphics[width=\textwidth]{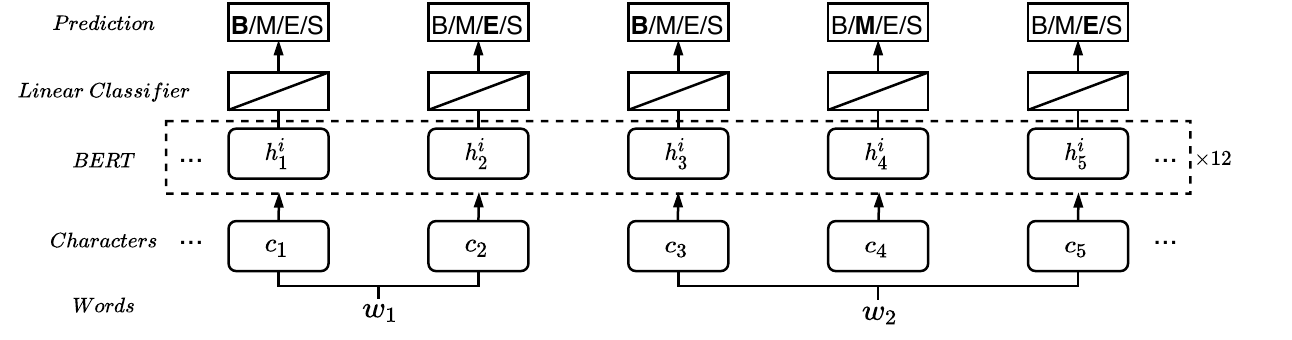}
	\caption{Word segmentation probing model.}
	\label{figure:probe}
\end{figure}

We probe the contextualized representation of each character for Chinese Word Segmentation (CWS)~\cite{cws}. In particular, CWS can be treated as a character-level sequence labeling task, where the label set includes \textsc{B}, \textsc{M}, \textsc{E}, and \textsc{S} (which stand for the beginning, middle, ending of word and single character word, respectively).  We directly use the fixed hidden representations in each layer as the input, on which a trainable linear classifier is built, as Figure~\ref{figure:probe} shows. We use local classifier rather than the conditional random fields~\cite{crf}, in order to focus on information extracted from hidden representations directly. The intuition is that if a simple linear classifier can predict the segmentation labels, we can reasonably conclude that the model has word structure features.  

\begin{table}[t]
	\centering
	\small
	\begin{tabular}{c|c|c|c|c|c|c|c|c|c}
		\hline 
		\multirow{2}{*}{\bf{Data}}&  \multicolumn{3}{c|}{\bf{Train}} &  \multicolumn{3}{c|}{\bf{Dev}} & 
		\multicolumn{3}{c}{\bf{Test}} \\ 
		\cline{2-10}
		& \#Sent  &\#Word &\#Char & \#Sent &\#Word  &\#Char & \#Sent &\#Word  &\#Char \\
		\hline
		CTB9 &  93.5K& 1.68M & 2.62M&9.28K & 142K & 217K&13.0K & 230K & 361K\\	
		\hline
		PKU &  17.1K & 1.01M & 1.66M &1.90K & 99.8K & 163K &1.94K & 104K & 172K\\
		\hline
		MSR & 78.2K & 2.12M & 3.63M & 8.69K & 246K & 417K & 3.98K & 106K & 184K	\\
		\hline
		CITYU & 47.7K & 1.29M & 2.14M  & 5.30K & 158K & 258K  & 1.49K & 40.9K & 67.6K\\
		\hline
		AS & 638K & 4.88M & 7.49M &70.8K & 563K & 876K &14.4K & 122K & 197K\\
		\hline
	\end{tabular}
	\caption{Statistics of datasets. }
	\label{table:statics}
\end{table}

Formally, in the representations of layer $l$, the label probability distribution of the character $c_j$ is calculated as follows:
\begin{equation}
P(y\prime|h^l_j) = \it{softmax}{(Wh_j^l)}
\end{equation}
where $y\prime$ $\in$ \{B, M, E, S\}, $W$ is the linear transformation parameters and $h_j^l$ is the $j$-th hidden representation in $l$-th layer.

%\footnote{\url{https://catalog.ldc.upenn.edu/LDC2016T13}}
\subsection{CWS Settings and Results}
We experiment with the CTB 9.0 dataset\footnote{\url{https://catalog.ldc.upenn.edu/LDC2016T13}}, and widely-used SIGHAN 2005 benchmarks\footnote{\url{http://sighan.cs.uchicago.edu/bakeoff2005/}} including four subsets (i.e, PKU, MSR, CITYU and AS). We split the CTB 9.0 dataset into train/dev/test set following Shao et al.~\shortcite{shao}. Statistics of the datasets are shown in Table~\ref{table:statics}. 
The standard word segmentation F1 score is used for evaluation. Our model is implemented using NCRF++~\cite{ncrfpp}. In particular, we use Adam~\cite{adam} as our optimization method, with a learning rate of 2e-5, a dropout rate of 0.1, and the number of training epochs being 3. The parameters are selected using the development set. 

Table~\ref{table:probeacc} shows the main results. %The bottom layer performs poorly compared with the other layers, which may be because the lowest layer encodes more about the character surface features. The performance is much improved for the second and third layers, which indicates that the model encode more word features. The middle layers (4$\sim$7) give the best results. For example, the layer 4 gives 92.20 F1-value in CTB9. For higher layers, the performance decreases, which may be because higher layers encode more semantic information beyond the word level.
In general, layers 3$\sim$8 give relatively higher performances, with F1 score close to or over 90\% for all the datasets. The best-performing layers are layers 4 and 7. Existing state-of-the-art neural segmentors trained on the datasets give F1 score of 96.5, 96.1, 97.4, 97.2, and 96.2  on CTB9, PKU, MSR, CITYU and AS datasets, respectively~\cite{shao-EtAl:2017:I17-1,glyce}. From the above results we can see that word information is strongly captured by BERT in the middle layers. This is consistent with findings of Section 3.

\iffalse
\begin{table}[t]
	\centering
	\begin{tabular}{c|c|c|c|c|c}
		\hline 
		\textbf{Layers} & \textbf{CTB9}  & \textbf{PKU}  & \textbf{MSR}  & \textbf{CITYU} & \textbf{AS}\\
		\hline
		\#1 &80.70&75.70&79.01&75.38 &78.75\\
		\#2 &88.91&85.21&86.31&86.33&87.73\\
		\#3 &91.54&88.43&88.66&90.57&91.24\\		
		\#4 &\textbf{92.20}&\textbf{88.74}&89.37&\textbf{91.26}&91.93\\
		\#5 &92.14&88.53&89.42&90.87&91.89\\
		\#6 &92.00&88.36&89.56&90.80&91.93\\
		\#7 &92.01&88.68&\textbf{89.69}&91.05&\textbf{92.04}\\
		\#8 &91.91&88.59&89.40&91.03&91.97\\
		\#9 &91.65&88.29&89.22&90.84&91.58\\
		\#10 &91.19&88.05&89.07&90.44&91.14\\
		\#11 &90.97&87.75&88.84&89.86&90.81\\
		\#12 &89.68&85.80&87.40&88.20&89.72\\
		%\cdashline{1-6}
		%\cite{shao} &96.67 &-&-&-&-\\
		%\cite{segdepqiu} &97.63 &-&-&-&-\\
		%\cite{glyce} &- &96.7&98.3&97.9&96.7\\
		\hline
	\end{tabular}
	\captionof{table}{Word segmentation results of our probing model.}
	\label{table:probeacc}
\end{table}
\fi

\begin{table}[t]
	\centering
	\small
	\begin{tabular}{c|c|c|c|c|c||c|c|c|c|c|c}
		\hline 
		\textbf{Layers} & \textbf{CTB9}  & \textbf{PKU}  & \textbf{MSR}  & \textbf{CITYU} & \textbf{AS} &\textbf{Layers} & \textbf{CTB9}  & \textbf{PKU}  & \textbf{MSR}  & \textbf{CITYU} & \textbf{AS}\\
		\hline
		\#1 &80.70&75.70&79.01&75.38 &78.75&\#7 &92.01&88.68&\textbf{89.69}&91.05&\textbf{92.04}\\
		\#2 &88.91&85.21&86.31&86.33&87.73&\#8 &91.91&88.59&89.40&91.03&91.97\\
		\#3 &91.54&88.43&88.66&90.57&91.24&	\#9 &91.65&88.29&89.22&90.84&91.58\\	
		\#4 &\textbf{92.20}&\textbf{88.74}&89.37&\textbf{91.26}&91.93&\#10 &91.19&88.05&89.07&90.44&91.14\\
		\#5 &92.14&88.53&89.42&90.87&91.89&\#11 &90.97&87.75&88.84&89.86&90.81\\
		\#6 &92.00&88.36&89.56&90.80&91.93&\#12 &89.68&85.80&87.40&88.20&89.72\\
		%\cdashline{1-6}
		%\cite{shao} &96.67 &-&-&-&-\\
		%\cite{segdepqiu} &97.63 &-&-&-&-\\
		%\cite{glyce} &- &96.7&98.3&97.9&96.7\\
		\hline
	\end{tabular}
	\captionof{table}{Word segmentation results of our probing model.}
	\label{table:probeacc}
\end{table}

\begin{table}[t]
	\centering
	\small
	\begin{tabular}{c|c|c|c|c|c|c}
		\hline 
		\textbf{Models (Best Layer)} & \textbf{CTB9}  & \textbf{PKU}  & \textbf{MSR}  & \textbf{CITYU} & \textbf{AS}& $\Delta$ (\textit{Avg.})\\
		\hline
		BERT-base &  92.20 &88.74&89.69&91.26&92.04 & -\\
		\hline 
		NER-finetune & 92.37  & 89.22&89.93&91.49& 92.25& $\Uparrow$ 0.27\\
		Chunking-finetune & 94.43   & 91.01&90.45&93.15& 93.74&$\Uparrow$ 1.77\\
		POS Tagging-finetune & \textbf{94.60}   & \textbf{91.16}&\textbf{90.61}&\textbf{93.42}& \textbf{93.86}&$\Uparrow$ 1.94\\
		
		CMNLI-finetune & 90.67   & 88.64&87.91&89.69&90.59&  $\Downarrow$ 1.28\\
		
		TNEWS-finetune & 92.05   & 88.57&89.15&90.99&91.92 &  $\Downarrow$ 0.25\\
		
		AFQMC-finetune & 92.04   & 88.85&89.14&91.30&92.04 &  $\Downarrow$ 0.11\\
		
		WSC-finetune & 92.14   &88.82 &89.30&91.33& 92.03&  $\Downarrow$ 0.06\\
		
		CSL-finetune &  92.17  & 88.80&89.29&91.37&92.17& $\Downarrow$ 0.02\\
		
		\hline
	\end{tabular}
	\captionof{table}{Performance of the word segmentation probing task after model finetuning.}
	\label{table:probeft}
\end{table}
\subsection{Tuning on Downstream Tasks}
To further analyze the influence of downstream tasks to word segmentation by fine-tuning the model using different downstream tasks including NER, chunking, POS tagging and five tasks in the Chinese Language Understanding Evaluation (CLUE) benchmark~\cite{clue}, then we execute the probing task again. Intuitively, word information is encoded by downstream tasks if the fine-tuned model performs better on word segmentation probing task, or vice versa.
The tasks and datasets we use are as follows:

%\noindent \textbf{Sentence Classification.}

%\noindent $\bullet\ \bf{IFLYTEK.}$ IFLYTEK is a long sentence classification dataset which is to
%assign each description into one of 119 categories, such as food, car rental, education, etc.

\noindent $\bullet\ \bf{NER.}$ We use the OntoNotes 4.0~\cite{weischedel2011ontonotes} as the named entity recognition dataset.

\noindent $\bullet\ \bf{Chunking.}$ CTB 4.0 is used as the chunking dataset. We split the data into train/dev/test set following Lyu et al.~\shortcite{Lyu:2016}.

\noindent $\bullet\ \bf{POS\ Tagging.}$ CTB 5.0 is used as the part-of-speech tagging dataset. The dataset split is the same as in Shao et al.~\shortcite{shao}.

\noindent $\bullet\ \bf{CMNLI}$ (Chinese Multi-Genre of Natural Language Inference) is a CLUE task to predict the relationship (neutral, entailment or contradiction) between two sentences.

\noindent $\bullet\ \bf{TNEWS}$ is a short sentence classification dataset\footnote{\url{https://github.com/fatecbf/toutiao-text-classfication-dataset/}} in CLUE, where the task is to assign a title to each news. The category of labels includes finance, technology, sports, etc.

\noindent $\bullet\ \bf{AFQMC}$ (Ant Financial Question Matching Corpus) comes from Ant Technology Exploration Conference (ATEC) Developer competition\footnote{\url{https://dc.cloud.alipay.com/index#/topic/intro?id=3}}. The CLUE task is to predict whether two sentences are semantically similar.

\noindent $\bullet\ \bf{WSC}$ (Winograd Schema Challenge)~\cite{wsc} Chinese datastet is a co-reference resolution task in CLUE.

\noindent $\bullet\ \bf{CSL}$ (Chinese Scientific Literature) dataset\footnote{\url{https://github.com/CLUEbenchmark/LightLM}} contains Chinese paper abstracts and their keywords. The CLUE task is to recognize whether given keywords are correct to the corresponding paper, where some noise keywords are generated by using TF-IDF value.

The results are shown in Table~\ref{table:probeft}. 
%The average performance of the word segmentation probing task slightly decreases after fine-tuning on task-specific dataset, which indicates word information is not necessary directly useful for these tasks, or at least not enhanced. CMNLI, TNEWS, and AFQMC are sentence-level classification tasks, which do not rely on the fine-grained word structure information. For WSC and CSL, the negative influence becomes smaller compared with the previous three tasks, this may because these tasks need some word information such as co-references word or keywords during training.
%The best fine-tuned models improve the result form 0.01 to 0.55 F1-value, which indicates the word information can be further detected during the fine-tuning from downstream tasks.
%One exception is NER, which is a word-sensitive sequence labeling task. The average F1-value increases by 0.27. One reasonable explanation is that word segmentation information is necessary for segmenting name entity~\cite{lattice-lstm}. During fine-tuning, word structure features can be further captured. 
%For NER, chunking and POS tagging, which are word-sensitive sequence labeling task The average F1-value increases by 0.27$\sim$1.94. One reasonable explanation is that word segmentation information is necessary and highly related to name entity, chunk and part-of-speech. During fine-tuning, word structure features can be further captured. 
We observe 5 trends for different datasets. First, for POS tagging and chunking, CWS probing is strongly improved after fine-tuning. These two datasets are the mostly connected with words, and joint segmentation models have been investigated~\cite{zhang-clark-2010-fast,zheng-etal-2013-deep,Lyu:2016}. Second, for NER, the performance slightly improved. The task sees debates on whether word information is useful, where segmentation error can outweight word features~\cite{debate2,debate1}. Third, for WSC and CSL the performance did not change. Fourth, for the text classification tasks TNEWS and AFQMC the performance slightly decreases. This may because word features do not help much for these tasks. Last, the performance decreases sharply for CMNLI, which shows that the semantics heavy task does not make use of word information and also worsen the performance of CWS probing model.

\begin{figure}[!t]
	\centering
	\includegraphics[width=\textwidth]{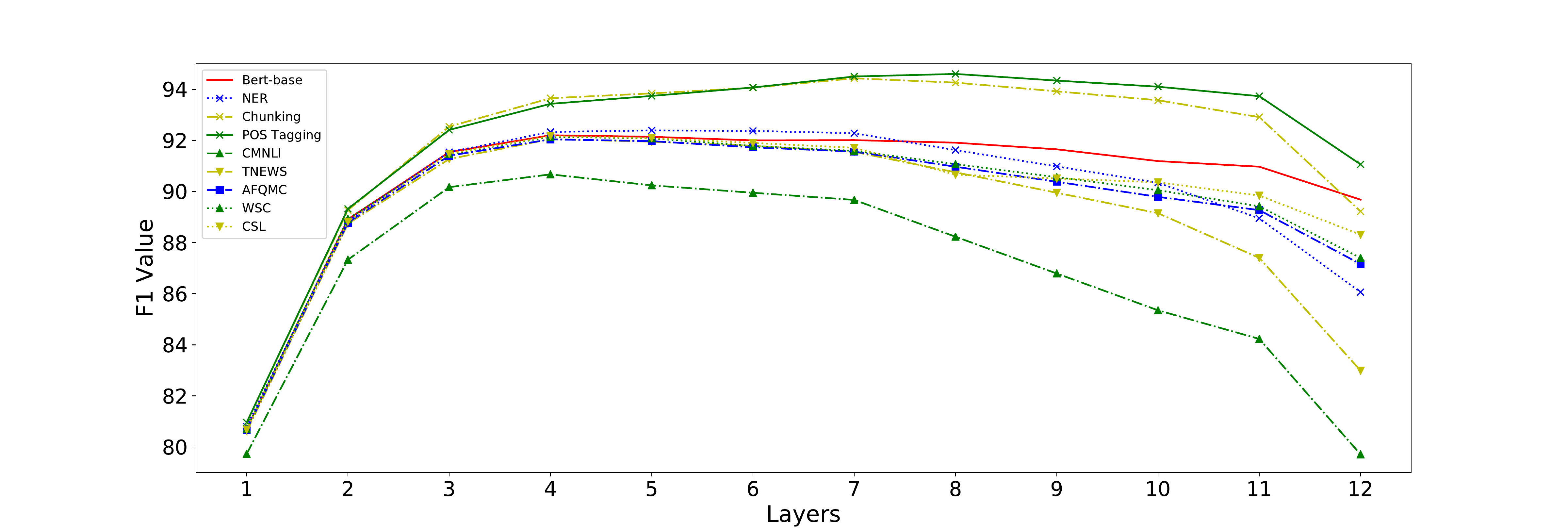}
	\caption{Layerwise word segmentation performance for BERT and fine-tuned models.}
	\label{figure:layers}
\end{figure}

We show the layer-wise word segmentation results on CTB9 for different models in Figure~\ref{figure:layers}. A silent finding is that for those tasks that benefit from word information, the best-performing layers move up from the middle layers. In contrast, for the CMNLI task, the best-performing layers move down from the middle layers. This shows that useful information can be brought closer to the final prediction layer during fine-tuning, or that training signals influence the top layers more strongly. For the other tasks, the best-performing layers are still 3$\sim$8.

%However, not all tasks benefit for the final results. For example, the performance on CITYU dataset does not get improvement , except a few from the CSL-finetuned model. The number of best layer stay unchanged or decrease by 1 or 2 layers, for example, the best layer \#11 in BERT-base models change to \#10 after fine-tuned in AFQMC task. This shows that the hierarchical features stays relatively stable, but it still can be affected by different downstream tasks.

%\subsection{Comparison of Models.}

%\textbf{BERT-base-wwm}

%TODO.
%Besides BERT-base-Chinese, \cite{cuiwwm} proposed

%we also select other variant of BERT for comparison: 

%\textbf{BERT-base-wwm} and RoBERTa-wwm-ext. These models mainly used the whole word masking technique for training objective and using external training corpus.\footnote{Pre-trained models downloaded from \url{https://github.com/ymcui/Chinese-BERT-wwm}} Results are shown in Table~\ref{table:probemodels}. 

\section{Related Work}
% \noindent \textbf{Find Knowledge from Pre-trained Models.} 
% \cite{elmo} find the pre-trained bidirectional LSTM model could encode syntactic and semantic features from different layers and performs well on correspond tasks. For models based on Transformer~\cite{transformer}, \cite{analysis-bert-1} also discover the BERT encodes surface, syntactic and semantic features hierarchically. \cite{petroni-etal-2019-language} find BERT contain some factual and commonsense knowledge on par with some existing structural databases. Few of work analysis the linguistic or structural knowledge from pre-trained model based on Chinese. 
\noindent \textbf{Knowledge from Pre-trained Models.} Peters et al.~\shortcite{elmo} find that ELMo encodes syntactic and semantic features at different layers. To analyze the syntactic features inside BERT, Goldberg~\shortcite{Goldberg2019AssessingBS} use BERT to predict the masked verb, and compare assigned scores between the correct verb and the incorrect verb. Petroni et al.~\shortcite{petroni-etal-2019-language} demonstrate that BERT is able to recall factual knowledge using pre-defined cloze templates. While all the above work has been conducted on English, little work analyzes the linguistic or structural knowledge from Chinese pre-trained model. We thus fill a gap in the literature.
%contain some factual and commonsense knowledge on par with some existing structural databases.

%For attention-based analysis,
%\noindent \textbf{Attention Analysis.} 
\noindent \textbf{Attention Analysis.} Clark et al.~\shortcite{attvis} visualize the attention patterns from BERT, finding different behaviors from different attention heads. 
Htut et al.~\shortcite{htut2019attention} evaluate syntactic knowledge by computing the maximum spanning tree on BERT's attention to recover dependency trees.
Kovaleva et al.~\shortcite{attpatterns} analyze the attention distribution for BERT fine-tuned across different tasks, pointing out that redundancy exists in different heads. Our work is similar in finding the patterns by making use of attention and fine-tuning. However, except token-to-token attention patterns, we also investigate attentions to specific characters for finding the potential word structure of Chinese.

\noindent \textbf{Probing Method.} Conneau et al.~\shortcite{probe1} introduce 10 probing tasks, uncovering linguistic properties that a sentence encoder captures. Liu et al.~\shortcite{probe2} design probing tasks on the contextualized representations from pre-trained models and investigate the linguistic knowledge it encodes. Ian et al.~\shortcite{tenney} design edge probing tasks to investigate the sub-sentential structure of contextualized
word embeddings. Hewitt and Manning~\shortcite{probe3syntax} propose structural probing methods and find that syntax trees are embedded implicitly in ELMo and BERT. Our method is similar in analyzing the contextualized representations directly. However, we select the Chinese word segmentation as our probing task for each character-level contextualized representation.

\section{Conclusion}
We investigated the capability of Chinese BERT for capturing the word structure using two different methods. First, analyzing the attention distribution for different patterns, we find that some of the attention heads can capture the word structure implicitly. Second, using a word segmentation probing task for the contextualized representation inside the model, we find that a simple linear classifier performs well in the middle layers. By using our probing method, we find evidence that different Chinese tasks rely on different degrees of word information, with NLI relying the least on word features.

% \section*{Acknowledgements}
% We thank all the anonymous reviewers for their thoughtful comments and suggestions. We gratefully acknowledge
% funding from the National Natural Science Foundation of China (NSFC No.61976180) and a research grant from Rxhui Inc. Yue Zhang is the corresponding author.

\bibliographystyle{coling}
\bibliography{coling2020}

\end{document}